\documentclass[journal]{IEEEtran}
\usepackage{amsmath,amsfonts}
\usepackage{algorithmic}
\usepackage{algorithm}
\usepackage{array}
\usepackage{textcomp}
\usepackage{stfloats}
\usepackage{url}
\usepackage{verbatim}
\usepackage{graphicx}
\usepackage{cite}

\usepackage[table,dvipsnames]{xcolor}
\usepackage{wrapfig}
\usepackage{subcaption}  
\usepackage{booktabs}
\usepackage{multirow}
\usepackage{rotating}
\usepackage[edges]{forest} % Loads forest, tikz, and the edges library
\usepackage{amsmath}
\usepackage{makecell}
\definecolor{coffee}{rgb}{0.44, 0.31, 0.22}
\begin{document}

\title{From Waveforms to Pixels: A Survey on Audio-Visual Segmentation}

\author{Jia Li, Yapeng Tian

\thanks{Jia Li, Department of Computer Science, The University of Texas at Dallas, Richardson, TX 75080, USA.}% 
\thanks{Yapeng Tian, Department of Computer Science, The University of Texas at Dallas, Richardson, TX 75080, USA.}
% \thanks{Manuscript received June 29, 2025.}
}

% The paper headers
\markboth{Journal of \LaTeX\ Class Files,~Vol.~14, No.~8, August~2021}%
{Jia Li, Yapeng Tian \MakeLowercase{\textit{et al.}}: From Waveforms to Pixels: A Survey on Audio-Visual Segmentation}

\maketitle

\begin{abstract}

Audio-Visual Segmentation (AVS) aims to identify and segment sound-producing objects in videos by leveraging both visual and audio modalities. It has emerged as a significant research area in multimodal perception, enabling fine-grained object-level understanding.
In this survey, we present a comprehensive overview of the AVS field, covering its problem formulation, benchmark datasets, evaluation metrics, and the progression of methodologies. We analyze a wide range of approaches, including architectures for unimodal and multimodal encoding, key strategies for audio-visual fusion, and various decoder designs. 
% Recent trends such as the adoption of transformer-based models and foundation models are also discussed. 
Furthermore, we examine major training paradigms, from fully supervised learning to weakly supervised and training-free methods. 
Notably, we provide an extensive comparison of AVS methods across standard benchmarks, highlighting the impact of different architectural choices, fusion strategies, and training paradigms on performance. 
Finally, we outline the current challenges, such as limited temporal modeling, modality bias toward vision, lack of robustness in complex environments, and high computational demands, and propose promising future directions, including improving temporal reasoning and multimodal fusion, leveraging foundation models for better generalization and few-shot learning, reducing reliance on labeled data through self- and weakly supervised learning, and incorporating higher-level reasoning for more intelligent AVS systems.
\end{abstract}


\vspace{-2mm}
\section{Introduction}
\label{sec:intro}

% .....................................................
Deep learning has significantly advanced visual segmentation, which involves identifying objects or regions and assigning pixel-level labels based solely on visual information~\cite{krizhevsky2012imagenet, long2015fully, chen2017deeplab, he2017maskrcnn, kirillov2019panoptic, xie2021segformer, kirillov2023segmentanything}. 
However, human perception is inherently multimodal, with audio and vision jointly contributing to a richer understanding of the environment~\cite{Chen1998Audio-visual, Frassinetti2002Enhancement, Nanay2017Multimodal, 10109890}. 
Audio-Visual Segmentation (AVS) has recently emerged as a fine-grained multimodal task aimed at identifying and segmenting the specific objects in a video that are responsible for producing sound, enabling dense scene understanding across modalities~\cite{zhou2022avs, liu2024bavs, wang2024gavs, gao2024avsegformer}. This capability holds great promise for real-world applications, ranging from advanced multimedia analysis and editing~\cite{mo2024multi, 7523378, 4202595}, to more intuitive human-computer interaction systems~\cite{yang2024analyzing, wang2024audiobench, johansen2022characterising, lv2022deep, fu2021design, 10387745}, and the development of robust autonomous systems capable of interpreting complex environments~\cite{schmidt2020acoustic, dutt2020self, topcu2020assured}.

Despite recent progress, AVS remains a challenging task due to several factors: 
(1) \textit{Modality heterogeneity}: Audio and visual signals differ fundamentally—audio is temporal or spectral, while visual data is spatial or spatiotemporal. These differences complicate feature alignment and fusion. (2) \textit{Precise localization}: Audio often captures scene-level context, while visual frames may contain multiple potential sources. Accurately localizing sound-producing pixels remains non-trivial. (3)\textit{ Potential temporal misalignments}: The duration of an audio event (e.g., a brief bark) may not perfectly align with the visual presence of the corresponding object (e.g., a dog remaining visible before and after barking). Such misalignments can lead to inaccurate audio-visual associations if not handled carefully. And lastly, (4) \textit{High annotation cost}: Annotating large-scale audio-visual datasets is expensive and complex, as it requires pixel-level, and temporally synchronized labels.

\begin{figure}[t]
    \centering
    \includegraphics[width=0.95\linewidth]{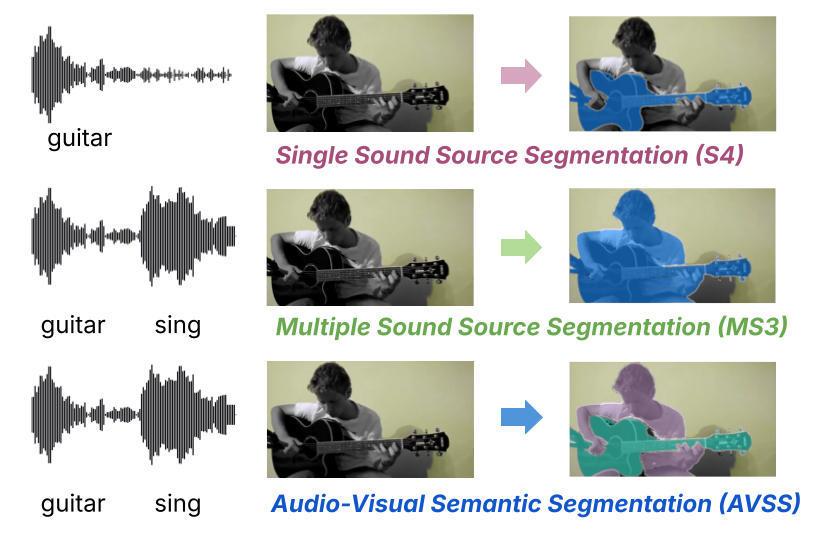}
    \caption{Illustration of three audio-visual segmentation (AVS) tasks with increasing complexity. {\textbf{Top}: Single Sound Source Segmentation} identifies and segments the primary sounding object in the scene. {\textbf{Middle}: Multiple Sound Source Segmentation} handles scenes with multiple active sound sources and segments all of them together. {\textbf{Bottom}: Audio-Visual Semantic Segmentation} not only segments the sounding objects but also assigns a semantic label (e.g., “guitar,” “singing”) to each one.}
    \label{fig:avs_trend_plot}
    \vspace{-6mm}
    
\end{figure}

Addressing these challenges relies on advancements in multimodal feature representation, cross-modal alignment techniques, and robust fusion strategies.
Since the introduction of its pioneering work in 2022~\cite{zhou2022avs}, the AVS field has developed rapidly. This growth is reflected in the rising number of publications, the creation of specialized datasets \cite{zhou2022avs, zhou2024avss, chen2024cavp-VPO}, and the diverse range of methods proposed to tackle core challenges~\cite{zhou2022avs, zhou2024avss, wang2024gavs, chen2024cavp-VPO, sun2023avs_false_neg, gao2024avsegformer}.
However, despite this progress, a dedicated survey analyzing recent AVS methodologies is missing. Existing related reviews~\cite{zhu2021deep, wei2022avc_survey,grumiaux2022ssl_survey}, while valuable, do not fill this specific niche. For instance, Wei et al.~\cite{wei2022avc_survey} explored the broader audio-visual learning context, categorizing various cross-modal tasks. While mentioning AVS, its wide scope prevented an in-depth discussion of AVS methods.
Another survey detailed deep learning for Sound Source Localization (SSL)~\cite{grumiaux2022ssl_survey}. However, that work largely predates current AVS advancements, and SSL, which typically provides coarse localization (e.g., bounding boxes or heatmaps), is distinct from AVS, which demands precise pixel-level segmentation masks (discussed further in Section~\ref{sec:task}).
% .....................................................
This survey aims to fill that gap by providing an in-depth, focused review of AVS—highlighting its key challenges, summarizing recent methods, and identifying promising future directions.
And to further guide research directions and benchmarking efforts, we present a detailed comparative analysis of representative AVS methods in Section~\ref{sec:results}. By examining their performance across standard datasets, we identify key factors influencing segmentation accuracy, such as the choice of encoders, fusion strategies, and training paradigms. 

The rest of the paper is organized as follows.
Section~\ref{sec:task} formally defines the AVS task and its objectives. Section~\ref{sec:methodologies} presents a comprehensive review of existing AVS methodologies, structured by architectural components (encoder, fusion, decoder) and training paradigms (supervised, weakly supervised, and training-free). Section~\ref{sec:benchmarks} introduces the commonly used datasets and evaluation metrics that facilitate empirical comparisons. Section~\ref{sec:results} analyzes the performance of representative models, drawing insights from benchmark results. Section~\ref{sec:challenges_future} outlines the key challenges that remain in the field and highlights promising directions for future research. Finally, Section~\ref{sec:conclusion} summarizes the survey’s main findings.

% \begin{table*}[t]
% \centering
% \caption{Comparison of related vision-language and audio-visual tasks.}
% \label{tab:task_comparison}
% \small
%     \begin{tabular}{lllll}
%         \toprule
%         \textbf{Task} & \textbf{Guidance Modality} & \textbf{Output} & \textbf{Annotation Type} & \textbf{Focus} \\
%         \midrule
%         AVS & Audio & Pixel-level mask & Segmentation mask & Sound-emitting objects \\
%         SSL & Audio & Heatmap & Bounding box & Approximate sound source \\
%         R-VOS & Text & Pixel-level mask & Segmentation mask & Text-described objects \\
%         \bottomrule
%     \end{tabular}
%     \vspace{-6mm}
    
% \end{table*}

\begin{table}[t]
\centering
\caption{Comparison of related vision-language and audio-visual tasks.}
\label{tab:task_comparison}
\footnotesize
\setlength{\tabcolsep}{3pt} % reduce column spacing
\begin{tabular}{lllll}
    \toprule
    \textbf{Task} & \textbf{Guidance} & \textbf{Output} & \textbf{Annotation} & \textbf{Focus} \\
    \midrule
    AVS & Audio & Pixel mask & Pixel mask & Sound-emitting obj. \\
    SSL & Audio & Heatmap & BBox & Approx. sound src. \\
    R-VOS & Text & Pixel mask & Pixel mask & Text-described obj. \\
    \bottomrule
\end{tabular}
\vspace{-3mm}
\end{table}

\vspace{-1mm}
\section{Task Definition and Formulation}
\label{sec:task}

\subsection{The problem}
\label{subsec:task_formulation}

AVS aims to identify and segment sounding objects within visual scenes~\cite{zhou2022avs,zhou2024avss,gao2024avsegformer}. Formally, given video frames \( \mathbf{V} \in \mathbb{R}^{T \times H \times W \times 3}\) and an audio waveform \(\mathbf{A} \), the overall objective is to learn a function \( f_\theta \), parameterized by \(\theta\), that maps the audio-visual input to a segmentation mask:
\vspace{-3mm}

\[
f_\theta : (\mathbf{V}, \mathbf{A}) \rightarrow \mathbf{M},
\]

\vspace{-3mm}
As illustrated in Figure~\ref{fig:avs_trend_plot}, the AVS task is categorized into several scenarios based on complexity. \textit{Single Sound Source Segmentation \underline{(S4)}} assumes one dominant sound source and produces a binary mask for it \(( \mathbf{M} \in {0,1}^{T \times H \times W} )\). In contrast, \textit{Multiple Sound Source Segmentation \underline{(MS3)}} handles multiple concurrent sources, generating a single unified mask for all sounding objects. 
A different challenge, addressed by tasks like \textit{\underline{VPO-MSMI}}, is to use spatial (e.g., binaural) audio to distinguish and segment multiple instances of the same object class, such as identifying which of two dogs is barking~\cite{chen2024cavp-VPO}.

Further extending the task, \textit{Audio-Visual Semantic Segmentation \underline{\textbf{}(AVSS)}} identifies the category of each source, outputting a per-pixel class-labeled mask \(\mathbf{M} \in {0, \dots, C}^{T \times H \times W} \) for a fixed set of \(C\) classes. A recent frontier, \textit{Open-Vocabulary AVS \underline{(OV-AVS)}}, generalizes this task to segment objects from novel semantic categories not seen during training~\cite{guo2024open}.

\vspace{-3mm}

\subsection{Related Tasks and Differences}
\label{subsec:related_tasks} % Added a label for the subsection

While AVS has been formally defined, it is essential to position it within the broader multimodal video understanding landscape. AVS shares components with tasks like Sound Source Localization (SSL) and Referring Video Object Segmentation (RVOS), but differs in goals and outputs. As summarized in Table~\ref{tab:task_comparison}, these tasks vary in their guiding modality, output format, and focus.

\paragraph{Sound Source Localization (SSL)}

SSL aims to identify the spatial origin of sounds within visual scenes, typically by predicting a bounding box or heatmap that highlights the most probable location of the sound-emitting source~\cite{arandjelovic2018objects, senocak2018learning, mo2022closer, chen2021localizing, mahmud2024t}. 
Unlike AVS, which generates fine-grained pixel-level object masks, SSL lacks instance-level segmentation and object-specific reasoning.

\paragraph{Referring Video Object Segmentation (R-VOS)}
R-VOS segments an object based on a natural language description~\cite{gavrilyuk2018rvos, botach2022rvos, wu2022language_vseg}. The key distinction is the guiding modality: R-VOS uses text to identify any referenced object, while AVS uses audio to segment only the object that is actively producing sound, focusing on audio-visual causality.

\vspace{-2mm}

\section{Methodologies}
\label{sec:methodologies}

\subsection{Architecture Overview}
\label{subsec:arch_overview}

\begin{figure}
    \centering
    \includegraphics[width=0.95\linewidth]{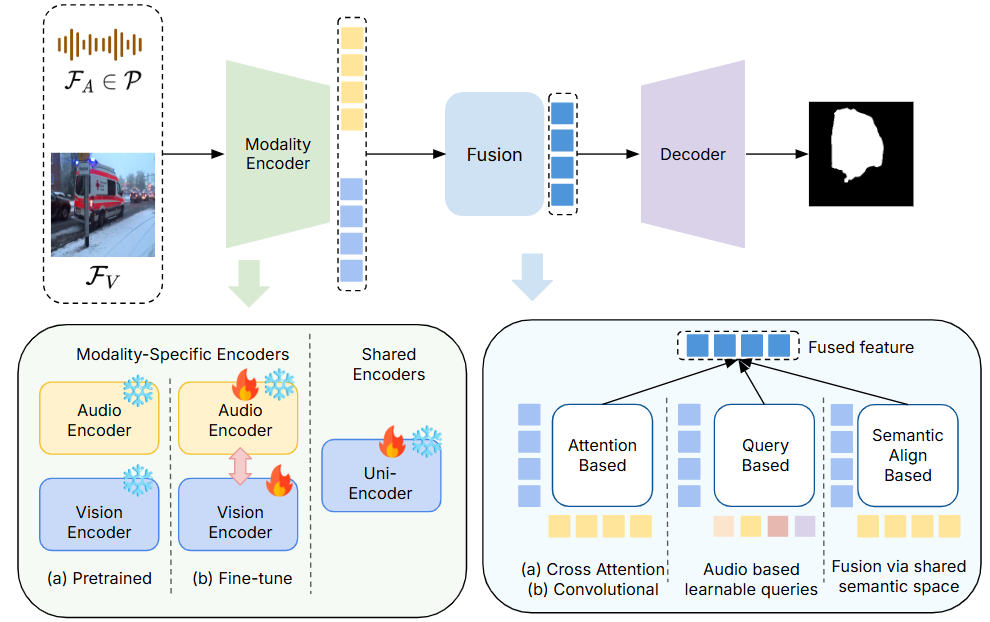}
    \caption{\textbf{Typical AVS architecture.} It includes modality encoders, a fusion module to connect different modalities, and a segmentation decoder. The encoders compress raw information, such as images or audio, into a more compact representation. There are broadly three types of fusion module: attention-based, query-based, and Semantic align-based. The decoder predicts segmentation masks based on the fused feature. }
    \label{fig:avs_arch}
    \vspace{-3mm}
\end{figure}

As illustrated in Figure \ref{fig:avs_arch}, most AVS models employ a three-stage pipeline that directly addresses the challenges described in Section \ref{sec:intro}. First, visual and audio encoders extract modality-specific features, mitigating heterogeneity. Then a fusion module using attention-, query- or semantic alignment–based strategies to integrates these features to align audio cues with image pixels and compensate for any temporal offsets. Finally, a segmentation decoder transforms the fused representation into a precise pixel-level mask. This streamlined architecture effectively tackles the primary challenges inherent in AVS, namely (1) modality heterogeneity, (2) fine-grained cross-modal alignment, and (3) temporal asynchrony.

To provide a clear overview, the following subsections will detail the methods used for these encoders (\S\ref{subsec:encoder}), fusion mechanisms (\S\ref{subsec:fusion}), and decoders (\S\ref{subsec:decoder}). Training strategies are discussed in \S\ref{subsec:training}.

\vspace{-2mm}

\subsection{Modality Encoders}
\label{subsec:encoder}

The encoder module converts raw audio-visual inputs into compact feature representations. AVS models typically adopt one of two architectural strategies: shared or modality-specific encoders.
\textit{A shared encoder} uses a single backbone (e.g., a Vision Transformer) for both audio spectrograms and visual frames. This approach prioritizes parameter efficiency but may struggle to capture unique modal characteristics~\cite{mo2023weakly-supervised, lin2023lavish}. A recent evolution of this strategy involves using unified multimodal foundation models like ImageBind~\cite{girdhar2023imagebind}, which are pretrained to produce inherently aligned cross-modal embeddings~\cite{girdhar2023imagebind, mo2024aligning}.
More commonly, models employ \textit{modality-specific encoders} to better preserve the distinct properties of each data stream. For vision, common backbones range from standard architectures like Swin Transformer~\cite{liu2021swin} to large foundation models such as CLIP~\cite{radford2021learning} and SAM~\cite{kirillov2023segmentanything}. For audio, prevalent choices include CNNs like VGGish~\cite{hershey2017vggish} and Transformers like HTS-AT~\cite{chen2022hts}, which are typically pretrained on large-scale audio datasets. Table~\ref{tab:encoders_refined} provides a detailed summary of these encoders.
To efficiently adapt these powerful pretrained models for AVS, a dominant approach is to freeze the encoder backbones and insert lightweight, trainable adapter modules. These modules, often using cross-attention, facilitate early audio-visual interaction and task-specific alignment without requiring costly full-model fine-tuning~\cite{lin2023lavish, liu2024SAMA-AVS-Syntheticdataset, duan2023DG-SCT}.

\begin{table}[t] % 
\small
\centering
\caption{Commonly used pre-trained encoders in AVS.}
\label{tab:encoders_refined} 
\scalebox{0.72}{
\begin{tabular}{lll}
\toprule
\textbf{Modality} & \textbf{Encoder Backbone} & \textbf{Typical Pre-training Dataset(s)} \\
\midrule
\multirow{5}{*}{Visual}
& ResNet~\cite{he2016resnet} & ImageNet~\cite{krizhevsky2012imagenet} \\
& PVT~\cite{wang2021pvt}  & ImageNet~\cite{krizhevsky2012imagenet} \\
& SwinTransformer~\cite{liu2021swin}  & ImageNet~\cite{krizhevsky2012imagenet} \\ % ImageNet is common initial pretraining
& CLIP~\cite{hafner2021clip}  & WebImageText (e.g., LAION-400M~\cite{schuhmann2021laion}) \\
& SAM~\cite{kirillov2023segmentanything}  & SA-1B~\cite{kirillov2023segmentanything} \\
\midrule
\multirow{4}{*}{Audio}
& VGGish~\cite{hershey2017vggish} & AudioSet~\cite{gemmeke2017audioset} \\
& PANNs~\cite{kong2020panns} & AudioSet~\cite{gemmeke2017audioset} \\
& BEATs~\cite{chen2022beats} & AudioSet~\cite{gemmeke2017audioset}, Speech Commands~\cite{warden2018speech}, etc. \\
& HTS-AT~\cite{chen2022hts} & AudioSet~\cite{gemmeke2017audioset}, Speech Commands~\cite{warden2018speech}, ESC-50~\cite{piczak2015esc} \\
& CLAP~\cite{elizalde2023clap} & LAION-Audio-630K~\cite{wu2023LAION}, AudioSet~\cite{gemmeke2017audioset}, etc. \\
\bottomrule
\end{tabular}
}

\vspace{-5mm}
\end{table}

\vspace{-2mm}
\subsection{Fusion Module: Integrating Audio and Visual Cues}
\label{subsec:fusion}

The fusion module lies at the core of AVS architectures, responsible for integrating the encoded audio and visual features. Its design directly addresses key AVS challenges: handling modality heterogeneity, enabling fine-grained spatial alignment, and managing temporal asynchrony. Fusion strategies are typically categorized by \textit{when} the modalities interact—during encoding (early fusion) or after encoding (late fusion)—and by \textit{how} the interaction is implemented (e.g., attention, queries, or alignment mechanisms).

\textbf{Early fusion} integrates audio-visual information during the encoding stage, typically through cross-modal interaction modules inserted into or between encoder layers, as discussed in \S\ref{subsec:encoder}. Methods such as LAVISH~\cite{lin2023lavish}, DG-SCT~\cite{duan2023DG-SCT}, and SAMA-AVS~\cite{liu2024SAMA-AVS-Syntheticdataset} adopt this strategy, enabling feature exchange and shared representation learning between modalities.

\textbf{Late fusion}, in contrast, combines audio and visual features (\(F_A, F_V\)) after they have been separately encoded. This strategy is widely adopted in AVS due to its modularity and simplicity. Interaction in late fusion is commonly achieved using \underline{\textit{cross-modal attention mechanisms}}~\cite{zhou2022avs, mo2023av_sam, liu2024SAMA-AVS-Syntheticdataset, mao2023ECMVAE, liu2024unlabel-frame, duan2023DG-SCT, chen2024cavp-VPO, mo2023weakly-supervised, yang2024combo, zhou2024avss}, which allow one modality to attend to the other; \underline{\textit{query-based methods}}~\cite{ma2024stepping, liu2024unlabel-frame, huang2023aqformer, liu2023autr, gao2024avsegformer, chen2024avsac, wang2024avesformer, wang2024gavs, shi2024c3n }, where learnable or semantic queries are used to extract relevant information from the fused streams; or \underline{\textit{alignment-based techniques}}~\cite{xu2024PIF, liu2024bavs, malard2024taco, bhosale2023leveraging, huang2024_rvos_gpt}, which project features into a shared space to facilitate correspondence and integration.

\subsubsection{Attention-Based Late Fusion.}
\label{parag:fusion_late_attn}

Attention-based late fusion methods primarily involve pixel-wise spatial attention~\cite{zhou2022avs, zhou2024avss}, channel-wise attention~\cite{gao2024avsegformer}, visual-to-audio cross-attention~\cite{chen2024cavp-VPO}, and bidirectional attention~\cite{yang2024combo}. Pixel-wise spatial attention~\cite{zhou2022avs, zhou2024avss} emphasizes relevant visual regions correlated with audio, while channel-wise attention selectively enhances entire feature channels based on audio cues~\cite{gao2024avsegformer}. Visual-to-audio cross-attention refines audio representations based on visual contexts~\cite{chen2024cavp-VPO}, and bidirectional attention enables mutual refinement of both modalities simultaneously~\cite{yang2024combo}.

\subsubsection{Query-Based Late Fusion.}
\label{parag:fusion_late_query}

Query-based late fusion methods build on the idea of integrating audio information into a set of learnable object queries, inspired by object detection frameworks like DETR~\cite{carion2020detr}. These queries—intended to represent sounding objects—are conditioned on audio features before attending to visual features via cross-attention in a transformer decoder. The core difference among methods lies in how they generate or refine these audio-aware queries. Some approaches use direct conditioning, where audio features are injected into or used to initialize queries (\textit{e.g.}, element-wise addition~\cite{li2023catr}, audio-initialized queries~\cite{liu2023autr}, or attention with audio as keys~\cite{huang2023aqformer}). Others adopt multi-stage refinement, such as combining audio-conditioned and learnable queries before decoding~\cite{gao2024avsegformer}. To enhance diversity and avoid overfitting to limited audio tokens, prompt-based generation methods concatenate audio features with learnable queries and refine them via self-attention~\cite{wang2024avesformer}, while adaptive selection strategies activate only queries most relevant to the input audio based on similarity with learned prototypes~\cite{ma2024stepping}. Additional methods apply semantic-aware modulation, where queries are filtered or enhanced based on semantic clusters derived from the audio~\cite{sun2023avs_false_neg} or by constructing enriched prompts using both audio and visual context~\cite{wang2024gavs}. CQFormer~\cite{lv2025cqformer} follows a similar late fusion paradigm but generates audio-visual queries through contrastive learning to improve cross-modal alignment and representation consistency.

\subsubsection{Alignment-Based Late Fusion.}
\label{parag:fusion_late_align}

Alignment-based late fusion focuses on improving audio-visual correspondence by explicitly aligning the two modalities—either at the semantic level or through data refinement—rather than directly fusing features via attention or queries. One approach achieves semantic consensus, where both audio and visual streams are processed through unimodal classifiers, and their predicted labels are embedded into a shared semantic space to identify a common concept guiding attention refinement (e.g., C3N \cite{shi2024c3n}). Another method emphasizes semantic correlation, where audio-predicted categories weight corresponding visual object masks to select semantically aligned regions (e.g., AVSC \cite{liu2023bias-mutual-semantics}). A distinct path is taken by data-level alignment, as in AVAgent \cite{mo2024aligning}, which uses an LLM-driven workflow—tool use, planning, and reflection—to iteratively refine the input audio-visual pair itself, correcting inconsistencies before model inference.

\vspace{-2.5mm}

\subsection{Decoder}
\label{subsec:decoder}

The decoder in an AVS model translates fused audio-visual features into final pixel-level masks by upsampling and refining spatial details. Early approaches often relied on convolution-based decoders inspired by architectures like U-Net~\cite{ronneberger2015u} and FPN~\cite{lin2017feature}, using transposed convolutions to generate masks~\cite{zhou2022avs}. More recently, Transformer-based decoders have become popular, leveraging audio-conditioned object queries refined via attention to capture global context and predict instance masks~\cite{huang2023aqformer, gao2024avsegformer}, with some variants using bidirectional fusion to address modality imbalance~\cite{chen2024avsac}. The latest trend involves foundation model-based decoders, which adapt large pretrained vision models like the Segment Anything Model (SAM)~\cite{kirillov2023segmentanything}. This is achieved either by using audio features to generate prompts for SAM’s mask decoder~\cite{mo2023av_sam, zhu2025ESAM} or by inserting lightweight adapters into its encoder for more efficient and deeper feature fusion~\cite{liu2024SAMA-AVS-Syntheticdataset}.

\begin{figure}[t] % Changed from [h] to [htbp]
  \centering
  \includegraphics[width=0.95\linewidth]{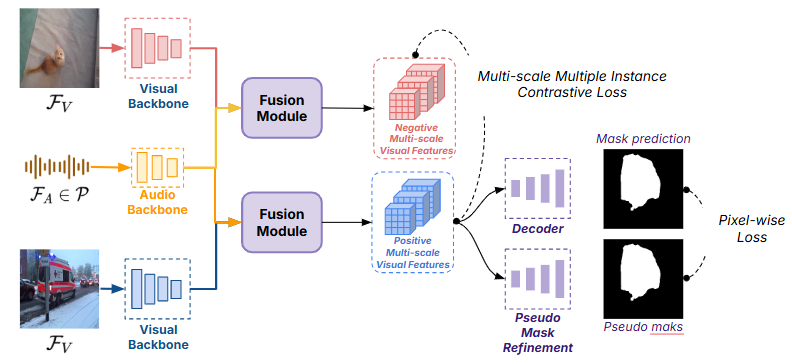}
  \caption{\textbf{Representative pipeline for weakly supervised audio-visual segmentation.} Audio and visual inputs are processed by separate backbones to extract global audio and multi-scale visual features. A fusion module aligns modalities via contrastive learning, while class-agnostic visual cues generate pseudo masks that supervise the decoder. The overall loss combines contrastive alignment and pixel-wise mask supervision. Adapted from WS-AVS~\cite{mo2023weakly-supervised}.}
  \label{fig:ws-avs}
    \vspace{-5mm}
  
\end{figure}

\begin{table*}[htbp]
\centering

\caption{Comparison of Audio-Visual Segmentation Datasets}
\label{tab:avs_datasets}

\resizebox{\textwidth}{!}{%
\begin{tabular}{lccccccccccc}
\toprule
\textbf{Dataset} & \textbf{Classes} & \textbf{Videos} & \textbf{Images} & \textbf{Masks} & \textbf{Bounding Box} & \textbf{Types} & \textbf{Annotations} & \textbf{Data Source} & \textbf{Human} & \textbf{Real-Data} & \textbf{Avg. Duration (s)} \\
\midrule
AVSBench - S4~\cite{zhou2022avs} & 23 & 4,932 & 24,660 & 10,852 & - & video & pixel & VGGSound~\cite{chen2020vggsound} & \checkmark & \checkmark & 5 \\
AVSBench - MS3~\cite{zhou2022avs} & 23 & 424 & 2,120 & 2,120 & - & video & pixel & VGGSound~\cite{chen2020vggsound} & \checkmark & \checkmark & 5 \\
AVSBench - semantic~\cite{zhou2024avss} & 70 & 12,356 & 82,972 & - & - & video & pixel \& category & VGGSound~\cite{chen2020vggsound} & \checkmark & \checkmark & 7.64 \\

AVS-Synthetic~\cite{liu2024SAMA-AVS-Syntheticdataset} & 46 & - & 58,405 & 62,609 & - & Image+Audio & pixel & LVIS~\cite{gupta2019lvis}; Open Images v4~\cite{kuznetsova2020openimagev4}; VGGSound~\cite{chen2020vggsound} & $\times$ & $\times$ & 3 \\
VPO-SS~\cite{chen2024cavp-VPO} & 21 & - & 12,202 & 12,202 & - & Image+Audio & pixel \& category & COCO~\cite{lin2014microsoft}; VGGSound~\cite{chen2020vggsound} & $\times$ & $\times$ & 10 \\
VPO-MS~\cite{chen2024cavp-VPO} & 21 & - & 9,817 & 13,496 & - & Image+Audio & pixel \& category & COCO~\cite{lin2014microsoft}; VGGSound~\cite{chen2020vggsound} & $\times$ & $\times$ & 10 \\
VPO-MSMI~\cite{chen2024cavp-VPO} & 21 & - & 12,855 & - & - & Image+Audio & pixel \& category \& instances & COCO~\cite{lin2014microsoft}; VGGSound~\cite{chen2020vggsound} & $\times$ & $\times$ & 10 \\

Robust\_AVSBench - S4~\cite{li2025reallyavs} & 23 & 4,932 & 24,660 & 10,852 & - & video & pixel & VGGSound~\cite{chen2020vggsound} & \checkmark & \checkmark & 5 \\
Robust\_AVSBench - MS3~\cite{li2025reallyavs} & 23 & 424 & 2,120 & 2,120 & - & video & pixel & VGGSound~\cite{chen2020vggsound} & \checkmark & \checkmark & 5 \\
% Robust\_AVSBench - semantic~\cite{li2025reallyavs} & 70 & 12,356 & 82,972 & - & - & video & pixel \& category & VGGSound~\cite{chen2020vggsound} & \checkmark & \checkmark & 7.64 \\

LU-AVS~\cite{liu2024LU-AVS} & 88 & 7,257* & 21,331,595 & 10,350,009 & 10,981,586 & video & pixel \& bounding box & VGGSound~\cite{chen2020vggsound} & \checkmark & \checkmark & 41.97 \\
\bottomrule
\multicolumn{12}{l}{\small *LU-AVS includes 6,627 normal videos and 630 hard videos}
\end{tabular}%
}

\vspace{-6mm}

\end{table*}

\vspace{-2mm}
\subsection{Learning Paradigm}
\label{subsec:training}

\subsubsection{Supervised and Semi-Supervised Learning}
\label{subsubsec:train_supervised}

This section summarizes the primary learning strategies in AVS, focusing on supervised and semi-supervised settings. While supervised learning uses full-frame annotations, semi-supervised variants often rely on only one annotated frame per clip. Both settings share similar training objectives, typically involving three main types of loss:

\textit{\underline{\textbf{Segmentation supervision losses}}} aim to directly improve mask quality. Common choices include Binary Cross-Entropy $\mathcal{L}_{\text{BCE}} = \text{BCE}(\mathbf{M}, \mathbf{Y})$)~\cite{zhou2022avs, lin2023lavish, mo2024aligning} 
and Dice loss (\(\mathcal{L}_{\text{Dice}} = 1 - (2 \sum \mathbf{M} \mathbf{Y} + 1) / (\sum \mathbf{M} + \sum \mathbf{Y} + 1)\) )~\cite{gao2024avsegformer}
, both of which compare the predicted mask $\mathbf{Y}$ with the ground truth $\mathbf{M}$ to optimize spatial accuracy.

\textit{\underline{\textbf{Cross-modal alignment losses}}} promote consistency between audio and visual features. These include KL divergence~\cite{zhou2022avs, zhou2024avss} to align feature distributions:
\[
\mathcal{L}_{\text{KL}} = \text{KL}( g(\mathbf{M} \odot \mathbf{V}_{\text{feat}}) || \mathbf{A}_{\text{feat}} ) 
\]
where \(\mathbf{M}\) is the predicted mask, \(\mathbf{V}_{\text{feat}}\) are visual features, \(\mathbf{A}_{\text{feat}}\) are audio features, and \(g(\cdot)\) is a global pooling function, 
and contrastive losses~\cite{chen2024cavp-VPO} 
which often achieved via an InfoNCE-style objective that distinguishes a matching audio-visual pair \((\mathbf{f}, \mathbf{f}^+)\) from a set of mismatched samples \(\mathcal{N}\):
\begin{equation}
    \mathcal{L}_{\text{contrastive}} = - \log \left( \frac{\exp(s^+)}{\exp(s^+) + \sum_{\mathbf{f}^- \in \mathcal{N}} \exp(s^-)} \right)
\end{equation}
where \(s^+ = \mathrm{sim}(\mathbf{f}, \mathbf{f}^+)/\tau\) and \(s^- = \mathrm{sim}(\mathbf{f}, \mathbf{f}^-)/\tau\) denote the scaled similarities for positive and negative pairs, respectively.

\textit{\underline{\textbf{Temporal consistency losses}}}~\cite{yang2024combo} ensure predictions remain stable across adjacent frames by penalizing dissimilarities between consecutive outputs. This is typically formulated as \(\mathcal{L}_{\text{temporal}} = \sum_{t=1}^{T-1} \mathcal{D}(\mathbf{M}_t, \mathbf{M}_{t+1})\), where \(\mathbf{M}_t\) is the prediction for frame \(t\) and \(\mathcal{D}\) is a dissimilarity metric (e.g., L1 distance or \(1-\cos(\cdot)\)).

In addition to standard end-to-end training, auxiliary strategies improve learning quality. \textit{Progressive learning}~\cite{ma2024stepping} separates training into stages, such as first localizing sounding objects and then refining semantics. \textit{Teacher-student frameworks}~\cite{liu2024unlabel-frame} extend supervision by generating pseudo-labels for unlabeled frames, leveraging temporal and motion cues to compensate for sparse annotations.

\subsubsection{Weakly Supervised Learning}
\label{subsubsec:train_weakly}

Weakly supervised learning (WSL)~\cite{zhou2018brief, li2015weakly, zhang2013representative, mo2023weakly-supervised} offers a scalable alternative to fully supervised AVS by using coarse labels—such as video-level tags instead of pixel-level annotations. The key challenge is translating weak global cues into dense segmentation masks.

A typical WSL pipeline (Figure~\ref{fig:ws-avs}), exemplified by methods like WS-AVS~\cite{mo2023weakly-supervised}, first uses contrastive learning to align audio-visual features, generating coarse spatial cues like attention maps. These noisy cues are improved through structured pseudo-label refinement to create higher-quality masks. These refined pseudo-masks then provide supervision to train a segmentation decoder, optimizing the entire framework end-to-end with a combined loss for both feature alignment and mask prediction.

\subsubsection{Training-free approaches}
\label{subsubsec:train_free}

Training-free methods~\cite{huang2024_rvos_gpt, bhosale2023leveraging, malard2024taco} bypass the need for model-specific AVS training by leveraging pre-trained foundation models~\cite{kirillov2023segmentanything, ravi2024sam, liu2024grounding} in modular pipelines. The core idea is to enable audio-guided segmentation by either translating audio cues into prompts for vision models or aligning audio and visual inputs within a shared embedding space.

One class of approaches focuses on \underline{\textit{prompt generation}}. For example, AL-Ref-SAM2~\cite{huang2024_rvos_gpt} uses GPT-4~\cite{achiam2023gpt} for reasoning over audio and video, generating language-based prompts to guide SAM2~\cite{ravi2024sam} for segmentation. Similarly, AT-GDINO-SAM~\cite{bhosale2023leveraging} converts audio into tags, detects relevant objects using Grounding DINO~\cite{liu2024grounding}, and segments them using SAM~\cite{kirillov2023segmentanything}.
Another class relies on \underline{\textit{embedding similarity}}. OWOD-BIND~\cite{bhosale2023leveraging} uses ImageBind~\cite{girdhar2023imagebind} to match object proposals and audio inputs in a joint embedding space. TACO~\cite{malard2024taco} aligns audio and visual features through a shared text space using CLAP~\cite{elizalde2023clap} and CLIP~\cite{hafner2021clip}, generating segmentations by identifying regions most semantically similar to the audio.

\vspace{-3mm}

\section{Datasets and Metrics}
\label{sec:benchmarks}

\vspace{-2mm}
\subsection{Datasets}
\label{subsec:datasets}

Progress in AVS is driven by dedicated datasets, which are broadly either real-world or synthetic (see Table~\ref{tab:avs_datasets} for a detailed comparison). Key \textit{real-world benchmarks} include AVSBench~\cite{zhou2022avs}, which provides single-source (S4) and multi-source (MS3) scenarios and was later given a semantic extension~\cite{zhou2024avss}, and LU-AVS~\cite{liu2024LU-AVS}, which offers long-form video annotations created via a semi-automatic pipeline~\cite{kirillov2023segmentanything, cheng2022xmem}. To overcome the high cost of manual annotation, various \textit{synthetic datasets} have been developed. These either compose samples from existing assets~\cite{liu2024SAMA-AVS-Syntheticdataset, gupta2019lvis}, use diffusion models for novel scene generation~\cite{senocak2024s3}, or are designed to test specific challenges, such as the progressively difficult settings in the VPO datasets~\cite{chen2024cavp-VPO} or the adversarial audio conditions for robustness evaluation in Robust-AVSBench~\cite{li2025reallyavs}.

\vspace{-3mm}

\subsection{Evaluation Metrics for AVS}
\label{subsec:metrics}

AVS performance is typically evaluated through spatial segmentation metrics, temporal localization metrics, and robustness metrics tailored to negative scenarios.

\textbf{Spatial metrics} focus on how well predicted masks align with ground truth at the frame level. The Jaccard Index (IoU) measures the overlap between predicted and ground truth masks, defined as \( J = \frac{|M_p \cap M_g|}{|M_p \cup M_g|} \), and is often averaged across frames (mIoU). The F-score, defined as the harmonic mean of precision and recall, emphasizes both accurate detection and suppression of false positives. In AVS, a weighted version with \( \beta^2 = 0.3 \) is commonly used to favor precision~\cite{zhou2022avs}.

\textbf{Temporal metrics}, such as temporal IoU (tIoU) and temporal F-score (tF), are critical for untrimmed video settings~\cite{liu2024LU-AVS}. The tIoU quantifies the overlap between predicted and ground truth time intervals of sound activity, while the tF-score evaluates the alignment of event boundaries via temporal precision and recall. 

\textbf{Robustness metrics} address model behavior under challenging conditions where no sounding object is present. For negative samples (e.g., silent or off-screen audio), the False Positive Rate (FPR) measures the fraction of pixels incorrectly classified as sounding objects. Since standard metrics (e.g. IoU and F-score) are ineffective when ground truth masks are empty, global versions G-mIoU and G-F are used to jointly evaluate accuracy on positive samples and suppression quality on negative ones~\cite{ssl_really, chen2024cavp-VPO}.

\begin{figure}[t]
    \centering
    \includegraphics[width=0.85\linewidth]{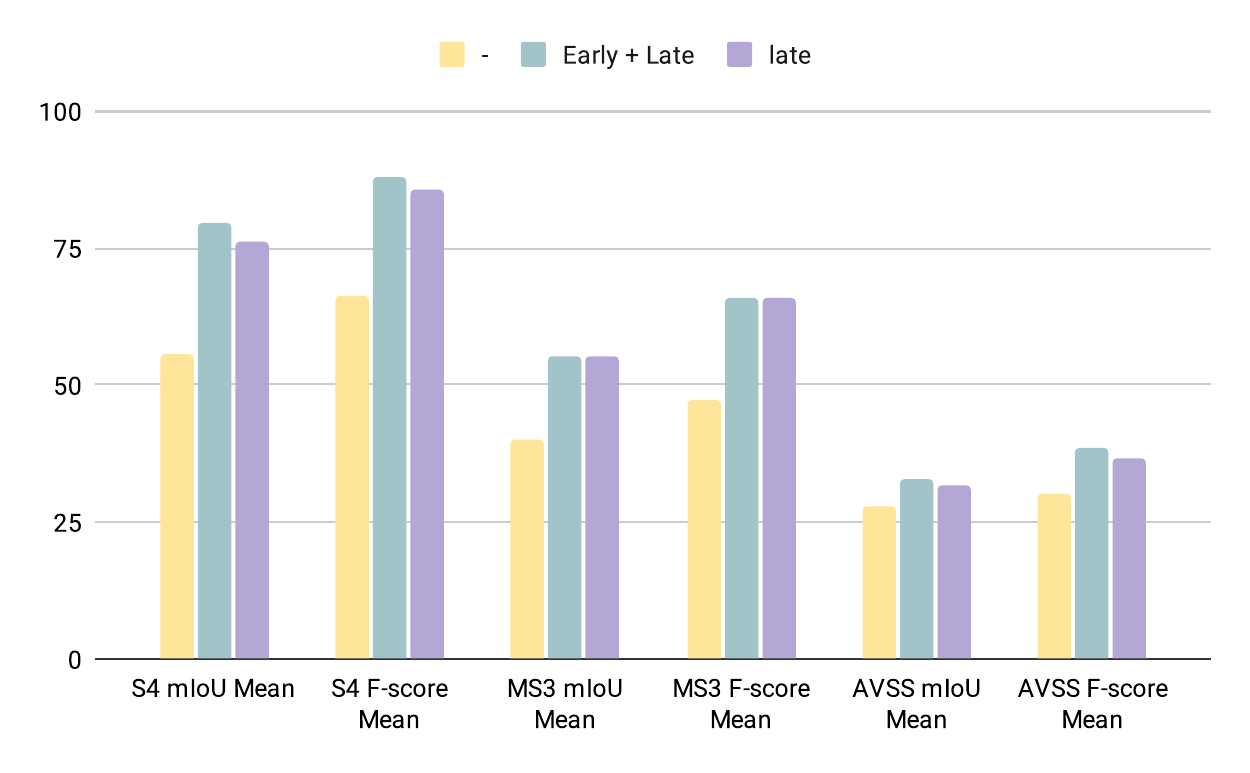}
    \vspace{-2mm}
    \caption{{Impact of Fusion Stage on AVS Performance.} Comparison of different fusion strategies on the S4, MS3, and AVSS benchmarks. Models combining both early and late fusion achieve the highest performance, followed by late-only fusion. Models with no explicit fusion perform significantly worse.}
    \label{fig:fusion-stage}

    \vspace{-5mm}
    
\end{figure}

\vspace{-2mm}
\section{Benchmarking and Findings}
\label{sec:results}

This section analyzes and discusses the performance of the various AVS methods reviewed in Section~\ref{sec:methodologies}. Our analysis primarily draws upon the comparative evaluation results presented in Table~\ref{tab:avsbench_result}, which summarizes the performance of representative models~\cite{zhou2022avs, mo2023av_sam, liu2024SAMA-AVS-Syntheticdataset, mao2023ECMVAE, liu2024unlabel-frame, duan2023DG-SCT, chen2024cavp-VPO, mo2023weakly-supervised, yang2024combo, zhou2024avss, ma2024stepping, liu2024unlabel-frame, huang2023aqformer, liu2023autr, gao2024avsegformer, chen2024avsac, wang2024avesformer, wang2024gavs, shi2024c3n , xu2024PIF, liu2024bavs, malard2024taco, bhosale2023leveraging, huang2024_rvos_gpt} on key benchmarks such as the AVSBench S4~\cite{zhou2022avs}, MS3~\cite{zhou2022avs}, and AVSS datasets~\cite{zhou2024avss}. As discussed below, performance can vary significantly based on architectural choices, training paradigms, and dataset characteristics. 

\begin{table*}[htbp]
% \begin{sidewaystable*}[t]

\centering
\rotatebox{90}{
\begin{minipage}{\textheight}
    \caption{Performance comparison of AVS methods across S4, MS3, and AVSS datasets.}
    \label{tab:avsbench_result}
    \scriptsize
    \begin{tabular}{llllllllrrllll}
    \toprule
     Models  & Reference& Visual Encoder& Audio Encoder& Training Approach & Fusion Stage& Fusion method& Decoder &S4 J &S4 F & MS3 J& MS3 F& AVSS J& AVSS F\\
    
     \midrule
     TPAVI& ECCV'22  & pvt  & VGGish  & (Semi) Supervised & late& Attention & FPN  & 78.7 &  87.9 & 54.0  & 64.5& 29.8 & 35.2 \\
     TPAVI& ECCV'22  & ResNet50 & VGGish  & (Semi) Supervised & late& Attention & FPN  & 72.8 &  84.8 & 47.9  & 57.8& 20.2 & 25.2 \\
     AQFormer & IJCAI'23 & pvt  & VGGish  & (Semi) Supervised & late& Query& Transformer& 81.6 &  89.4 & 61.1  & 72.1& - & -\\
     AQFormer & IJCAI'23 & ResNet50 & VGGish  & (Semi) Supervised & late& Query& Transformer& 77&  86.4 & 55.7  & 66.9& - & -\\
     AUTR & arXiv'23 & pvt  & VGGish  & (Semi) Supervised & late& Query& Transformer& 80.4 &  90& 56.2  & 70.0& - & -\\
     AUTR & arXiv'23 & ResNet50 & VGGish  & (Semi) Supervised & late& Query& Transformer& 75&  85.2 & 49.4  & 61.2& - & -\\
     AV-SAM  & arXiv'23 & SAM  & Resnet18  & (Semi) Supervised & late& Attention & SAM  & 40.5 &  56.6 & -& -  & - & -\\
     CATR & ACM MM'23& pvt  & VGGish  & (Semi) Supervised & early& Query& Transformer& 81.4 &  89.6 & 59.0  & 70.0& 32.8 & 38.5 \\
     CATR & ACM MM'23& ResNet50 & VGGish  & (Semi) Supervised & early& Query& Transformer& 74.8 &  86.6 & 52.8  & 65.3& - & -\\
     DIFFAVS & arXiv'23 & pvt  & VGGish  & (Semi) Supervised & late& Diffusion & FPN  & 81.4 &  90.2 & 58.2  & 70.9& - & -\\
     DIFFAVS & arXiv'23 & ResNet50 & VGGish  & (Semi) Supervised & late& Diffusion & FPN  & 75.8 &  86.9 & 49.8  & 58.2& - & -\\
     ECMVAE  & ICCV'23  & pvt  & VGGish  & (Semi) Supervised & early& Attention & FPN  & 81.7 &  90.1 & 57.8  & 70.8& - & -\\
     ECMVAE  & ICCV'23  & ResNet50 & VGGish  & (Semi) Supervised & early& Attention & FPN  & 76.3 &  86.5 & 48.7  & 60.7& - & -\\
     CMMS~\cite{liu2023bias-mutual-semantics}& ACM MM'23& ResNet50 & BEATs& (Semi) Supervised & early& Query& Transformer& 77&  85.2 & 49.6  & 61.5& - & -\\
     CMMS~\cite{liu2023bias-mutual-semantics}& ACM MM'23& pvt  & BEATs& (Semi) Supervised & early& Query& Transformer& 80.6 &  88.2 & 58.2  & 65.1&    &   \\
     SAM  & ICCV'23  & SAM  & -& Training-Free  & -  & -& SAM  & 55.1 &  73.9 & 54.0  & 63.9& - & -\\
     LAVISH  & CVPR'23  & vit-b/16 & vit-b/16  & (Semi) Supervised & early& Attention & FPN  & 80.1 &  88& 49.8  & 60.3& - & -\\
     DG-SCT  & Neurips'23  & Swin-v2-L  & HTS-AT  & (Semi) Supervised & early& Attention & FPN  & 80.9 &  89.2 & 53.5  & 64.2& - & -\\
     
    \midrule
     
     AP-SAM  & WACV'24  & SAM  & VGGish  & (Semi) Supervised & -  & Query& SAM  & 69.6 &  79.6 & 51.6  & 57.8& - & -\\
     AVESFormer\_18 & arXiv'24 & ResNet18 & VGGish  & (Semi) Supervised & late& Query& ELF  & 77.3 &  87.5 & 55.5  & 65.1& 26.3 & 31.8 \\
     AVESFormer\_50 & arXiv'24 & ResNet50 & VGGish  & (Semi) Supervised & late& Query& ELF  & 79.9 &  89.1 & 57.9  & 68.7& 31.2 & 36.8 \\
     AVSAC& TCSVT'24 & pvt  & VGGish  & (Semi) Supervised & late& Query& Dual Transformer & 84.5 &  91.6 & 64.2  & 76.6& 37.0 & 42.4 \\
     AVSAC& TCSVT'24 & ResNet50 & VGGish  & (Semi) Supervised & late& Query& Dual Transformer & 76.9 &  87& 54.0  & 65.8& 25.4 & 29.7 \\
     AVSBiGen & AAAI'24  & pvt  & VGGish  & (Semi) Supervised & late& Attention & FPN  & 81.7 &  90.4 & 55.1  & 66.8& - & -\\
     AVSBiGen & AAAI'24  & ResNet50 & VGGish  & (Semi) Supervised & late& Attention & FPN  & 74.1 &  85.4 & 45.0  & 56.8& - & -\\
     AVSegFormer& AAAI'24  & pvt  & VGGish  & (Semi) Supervised & late& Query& Transformer& 83.1 &  90.5 & 61.3  & 73.0& 37.3 & 42.8 \\
     AVSegFormer& AAAI'24  & ResNet50 & VGGish  & (Semi) Supervised & late& Query& Transformer& 76.4 &  86.7 & 53.8  & 65.6& 26.6 & 31.5 \\
     BAVS & TMM'24& pvt  & VGGish  & (Semi) Supervised & late& Alignment & Transformer& 82.7 &  89.8 & 59.6  & 65.9& 33.6 & 37.5 \\
     BAVS & TMM'24& ResNet50 & VGGish  & (Semi) Supervised & late& Alignment & Transformer& 78&  85.3 & 50.2  & 62.4& 24.7 & 29.6 \\
     CAVP & CVPR'24  & ResNet50 & VGGish  & (Semi) Supervised & late& Attention & FPN  & 78.7 &  88.8 & 45.8  & 61.7& - & -\\
     CLIP help  & WACV'24  & CLIP & BEATs& (Semi) Supervised & late& clip\_seg  & Transformer& 59.8 &  69& 41.1  & 46.7& - & -\\
     COMBO& CVPR'24  & pvt  & VGGish  & (Semi) Supervised & late& Attention & Transformer& 84.7 &  91.9 & 59.2  & 71.2& 42.1 & 46.1 \\
     COMBO& CVPR'24  & ResNet50 & VGGish  & (Semi) Supervised & late& Attention & Transformer& 81.7 &  90.1 & 54.5  & 66.6& 33.3 & 37.3 \\
     CPM  & ECCV'24  & ResNet50 & VGGish  & (Semi) Supervised & late& Alignment & Transformer& 81.4 &  90.5 & 59.8  & 71.0& 34.5 & 39.6 \\
     GAVS & AAAI‘24  & SAM  & VGGish  & (Semi) Supervised & late& Query& Transformer& 80.1 &  90.2 & 63.7  & 77.4&    &   \\
     PIF  & TMM'24& pvt  & VGGish  & (Semi) Supervised & late& Alignment & Transformer& 81.4 &  90& 58.9  & 70.9& - & -\\
     PIF  & TMM'24& ResNet50 & VGGish  & (Semi) Supervised & late& Alignment & FPN  & 75.4 &  86.1 & 53.9  & 65.4& - & -\\
     SAMA-AVS & WACV'24  & SAM  & VGGish  & (Semi) Supervised & early& Attention & SAM  & 81.5 &  88.6 & 63.1  & 69.1& - & -\\
     SeIM & ACM MM'24& ResNet50 & VGGish  & (Semi) Supervised & early& Mamba& Clever& 76.6 &  86.2 & 54.5  & 65.6& - & -\\
     SelM & ACM MM'24& pvt  & VGGish  & (Semi) Supervised & early& Mamba& Clever& 83.5 &  91.2 & 60.3  & 71.3& - & -\\
     Stepping-Stones  & ECCV'24  & pvt  & VGGish  & (Semi) Supervised & late& Query& Transformer& 83.2 &  91.3 & 67.3  & 77.6& - & -\\
     UFE\_AVSegformer & CVPR'24  & pvt  & VGGish  & (Semi) Supervised & late& Query& Transformer& 83.2 &  90.4 & 62.0  & 70.9& - & -\\
     UFE\_AVSegformer & CVPR'24  & ResNet50 & VGGish  & (Semi) Supervised & late& Attention & FPN  & 79&  87.5 & 55.9  & 64.5& - & -\\
     UFE\_TPAVI & CVPR'24  & pvt  & VGGish  & (Semi) Supervised & late& Query& Transformer& 82.5 &  90& 59.5  & 70.0& - & -\\
     UFE\_TPAVI & CVPR'24  & ResNet50 & VGGish  & (Semi) Supervised & late& Attention & FPN  & 78.2 &  88.7 & 54.1  & 61.6& - & -\\
     WS-AVS  & NeurIPS'24  & ResNet50 & ResNet50  & Weakly-Supervised & late& Attention & FPN  & 34.1 &  51.8 & 30.9  & 46.9& - & -\\
    
     TeSO\_swin  & ECCV'24  & Swin-B& BEATs&   &   &   &    & 83.3 &  93.3 & 66.0  & 80.1& 68.5 & 81.3 \\
     \midrule
     AL-Ref-SAM2& AAAI‘25  & GPT  & BEATs& Training-Free  & -  & Alignment & SAM  & 70.5 &  81.1 & 48.6  & 53.5& 36.0 & 39.8 \\
     AT-GDINO-SAM  & AAAI'25  & DINO & AST & Training-Free  & -  & Alignment & SAM  & 38&  46& 25.0  & 29.0& 24.0 & 25.0 \\
     AVAgent & NeurIPS'25  & Imagebind  & Imagebind & (Semi) Supervised &   &   &    & 36.4 &  49.9 & -& -  & - & -\\
     % AVS-Mamba  & arXiv'25 & pvt  & VGGish  & (Semi) Supervised &   &   &    & 85&  92.6 & 68.6  & 78.8& - & -&   &  &   &  &  &    \\
     % AVS-Mamba  & arXiv'25 & ResNet50 & VGGish  & (Semi) Supervised &   &   &    & 78.6 &  88.9 & 63.9  & 74.9& - & -&   &  &   &  &  &    \\
     MoCA & AAAI'25  & Imagebind  & Imagebind & Weakly-Supervised &   &   &    & 68&  79& 57.0  & 62.0& 31.0 & 33.0 \\
     OWOD-BIND  & AAAI'25  & OWOD & Imagebind & Training-Free  & -  & Alignment & SAM  & 58&  67& 34.0  & 44.0& - & -\\
     SAM-BIND & AAAI'25  & SAM  & Imagebind & Training-Free  & -  & Alignment & SAM  & 42&  51& 28.0  & 36.0& 24.0 & 26.0 \\
     % AVS-Mamba  & Arxiv'25 & ResNet50 & VGGish  & (Semi) Supervised & late& Mamba& Modality Aggregation Decoder & 78.6 &  88.9 & 63.9  & 74.9& 32.2 & 38.2 &   &  &   &  &  &    \\
     % AVS-Mamba  & Arxiv'25 & pvt  & VGGish  & (Semi) Supervised & late& Mamba& Modality Aggregation Decoder & 85&  92.6 & 68.6  & 78.8& 39.7 & 45.1 &   &  &   &  &  &    \\
     TACO & Arxiv'25 & CLIP & CLAP & Training-Free  & late& Alignment & FC-CLIP  & 64&  71.5 & 43.2  & 47.5&    &   \\
    \bottomrule
    \end{tabular}

    \end{minipage}
    }
\end{table*}
% \end{sidewaystable*}

\paragraph{Impact of Architectural Components - Encoder}

\textbf{\textit{Transformer-based visual encoders significantly outperform CNN-based alternatives in AVS tasks.}}
The choice of visual encoder heavily impacts AVS performance, as it determines the quality of features available for audio-visual alignment. As shown in Table~\ref{tab:avsbench_result}, models using Transformer-based backbones like Swin Transformer~\cite{liu2021swin} and PVT~\cite{wang2021pvt} consistently outperform those with conventional CNNs like ResNet~\cite{he2016resnet}. This advantage stems from their superior ability to capture the rich semantic context for accurately segmenting complex sound-producing objects.

\paragraph{Impact of Architectural Components - Fusion Stage}
\label{subsec:fusion_stage}

\textbf{\textit{Incorporating both early and late fusion stages leads to more robust and higher-performing AVS models.}}
The stage at which audio and visual features are fused significantly impacts model performance and robustness. As demonstrated in Table~\ref{tab:avsbench_result} and Figure~\ref{fig:fusion-stage}, architectures that combine both early and late fusion consistently achieve the highest segmentation scores and the lowest performance variance across all benchmarks. In contrast, strategies using only late fusion perform moderately worse and less consistently, while models with no explicit fusion mechanism perform the poorest. This pattern emphasizes that injecting audio information at multiple levels of the visual processing hierarchy is crucial for achieving effective and stable segmentation.

\paragraph{Impact of Architectural Components - Fusion Strategies}
\label{subsec:fusion_strategies}

\begin{figure}[t]
    \centering
    \includegraphics[width=0.85\linewidth]{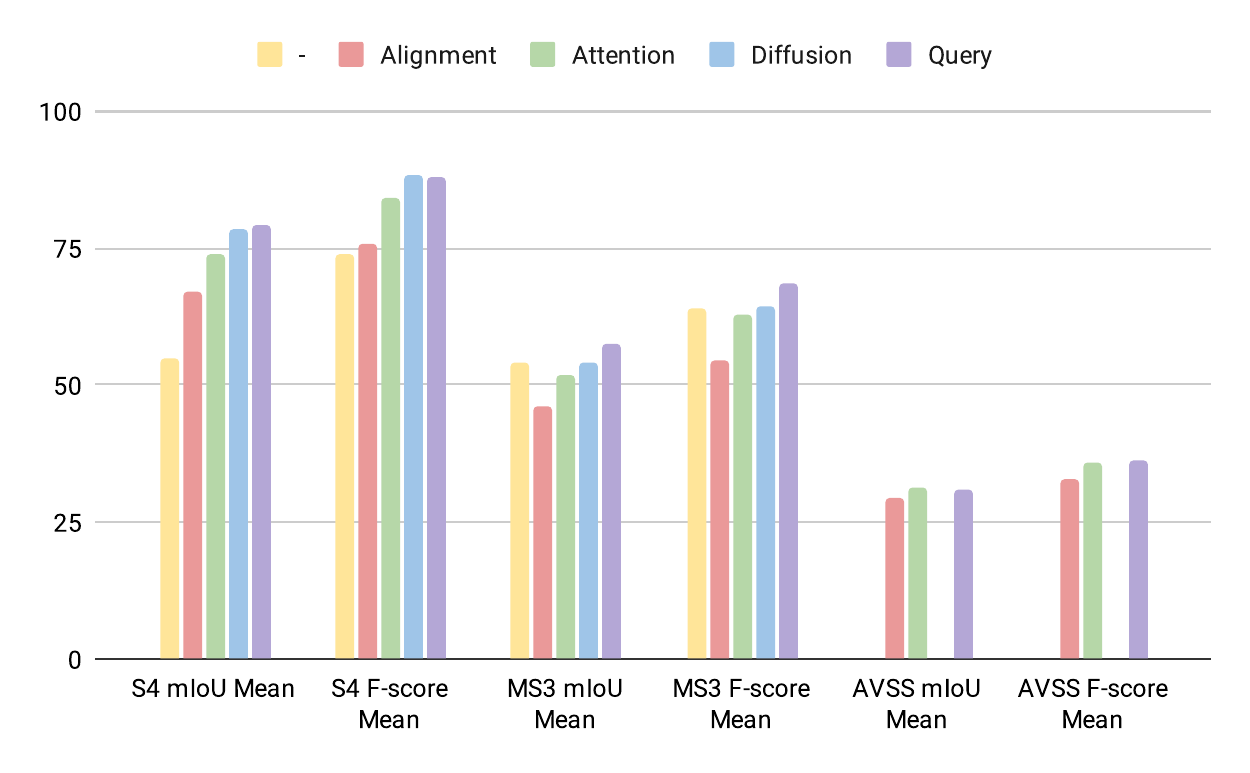}
    \vspace{-2mm}
    \caption{{Performance Comparison of AVS Fusion Strategies.} The chart evaluates different fusion methods across S4, MS3, and AVSS benchmarks. Query-based fusion consistently delivers the highest performance in both mIoU and F-score, outperforming diffusion, attention, and alignment-based approaches.}
    \label{fig:fusion-method}
    % \vspace{-3mm}
\end{figure}

\textbf{\textit{Query-based fusion strategies consistently deliver the best performance and stability across AVS benchmarks.}}
The fusion strategy is central to AVS, and as shown in Figure~\ref{fig:fusion-method} and Table~\ref{tab:avsbench_result}, a clear performance hierarchy exists. Query-based fusion consistently achieves the highest segmentation scores and the most stable performance across all benchmarks. Other approaches like attention-based fusion are also competitive, effectively capturing nuanced modality dependencies. In contrast, alignment-based strategies yield significantly lower and more variable results, suggesting less stable cross-modal integration.

\paragraph{Influence of Training Paradigms}

\textbf{\textit{Supervised training paradigms yield the highest AVS performance, while weakly supervised and training-free approaches offer trade-offs between accuracy and annotation efficiency.}}
The chosen training paradigm directly dictates this trade-off between performance and annotation cost. As detailed in Table~\ref{tab:avsbench_result}, semi- or fully supervised methods that fine-tune models on pixel-level annotations consistently achieve state-of-the-art performance across all metrics. In contrast, weakly supervised approaches~\cite{mo2023weakly-supervised}, which generate pseudo-masks to avoid dense annotation, offer a competitive but less accurate alternative. Meanwhile, training-free methods provide a zero-shot solution by composing large foundation models, further trading accuracy for maximum efficiency by eliminating task-specific fine-tuning.

\paragraph{Performance on LU-AVS Benchmark~\cite{liu2024LU-AVS}}

\textbf{\textit{Current AVS models struggle to localize sound-making objects in long untrimmed videos, with all methods performing poorly on the LU-AVS benchmark.}}
As shown in Table~\ref{tab:stag_comparison}, even state-of-the-art methods like TPAVI~\cite{zhou2022avs}, ECMVAE~\cite{mao2023ECMVAE}, and the benchmark's own baseline STAG~\cite{liu2024LU-AVS} perform poorly. Their mean temporal (m\_tIoU) and visual (m\_vIoU) IoU scores all fall below 20\%, underscoring the significant challenge posed by long-form video segmentation.

\begin{table}[t]
\centering

\caption{Performance comparison of different methods on the LU-AVS~\cite{liu2024LU-AVS} benchmark. Metrics shown are mean temporal/visual IoU and F-score.}
\label{tab:stag_comparison}

\resizebox{0.75\linewidth}{!}{%
\begin{tabular}{l|cccc}
\toprule
Method & m\_tIoU & m\_vIoU & m\_tF & m\_vF \\
\midrule
TPAVI~\cite{zhou2022avs} & 14.12 & 6.03 & 14.35 & 6.73 \\
ECMVAE~\cite{mao2023ECMVAE} & 13.01 & 5.24 & 13.86 & 5.37 \\
STAG (Mask)~\cite{liu2024LU-AVS} & 18.76 & 17.32 & 17.33 & 16.25 \\
STAG (BBox)~\cite{liu2024LU-AVS} & 16.53 & 15.89 & -- & -- \\
\bottomrule
\end{tabular}
}

\vspace{-5mm}

\end{table}

\begin{table*}[!htbp]
\centering

\caption{Performance comparison of various models on different audio input types and global metrics.}
\label{tab:robustAVS}

\resizebox{\textwidth}{!}{%
\begin{tabular}{c|c|ccc|cc|ccc|ccc|ccc|ccl}
\hline

& & \multicolumn{3}{c|}{{Global metric}}
& \multicolumn{2}{c|}{{Positive audio input}} & \multicolumn{9}{c|}{{Negative audio input}} 
& \multicolumn{3}{c}{{Unseen Audio}}\\

& & & 
& & \multicolumn{2}{c|}{}&  \multicolumn{3}{c}{{Slience}}&\multicolumn{3}{c}{{Noise}}
& \multicolumn{3}{c|}{{Offscreen sound}} & & &\\ \hline

{Test set} & {Model}& {G-mIoU↑} & {G-F↑} & {G-FPR↓} 
&{mIoU ↑}& {F-score ↑}& {mIoU ↓}&  {F-score ↓}&{FPR ↓}&{mIoU ↓}& {F-score ↓}& {FPR ↓}& {mIoU ↓}&  {F-score ↓}&{FPR ↓} 
& {mIoU ↓}&{F-score ↓} &{FPR ↓} \\ \hline

\multirow{7}{*}{{AVSBench-S4}}

& TPVAI~\cite{zhou2024avss}
& {35.032}& {21.479}& {0.186}
& {78.7}& {87.9}& {76.6}&    {87.1}&{18.74}
& {77.6}& {88.0}& {18.44}
&  {78.2}&   {88.2}&{18.66}
& 78.22& 87.52&18.69\\

& AVSegFormer~\cite{gao2024avsegformer}
& {28.199}& {17.355}& {0.188}
& {82.1}& {89.9}& {83.0}&    {90.4}&{18.81}
& {83.0}& {90.4}& {18.81}
&  {83.0}&   {90.4}&{18.81}
&  {82.98}&  {90.40}&{18.81}\\

& Stepping-Stones~\cite{ma2024stepping}
& {28.980}& {15.806}& {0.190}
& {83.2}& {91.3}& {82.2}&    {91.3}&{18.67}
& {82.2}& {91.3}& {19.05}
&  {82.5}&   {91.3}&{19.00}
& 82.51& 91.27&19.02\\

& SAMA-AVS~\cite{liu2024SAMA-AVS-Syntheticdataset}
& {52.688}& {40.417}& {0.155}
& {83.1}& {90.0}& {56.2}&    {69.1}&{17.26}
& {59.3}& {73.8}& {13.42}
&  {68.7}&   {79.0}&{17.07}
& {69.7}& {79.9}&{17.00}\\

& CAVP~\cite{chen2024cavp-VPO}
& {33.526}& {19.891}& {0.185} 
& {78.7}& {88.8}& {78.7}&    {88.8}&{18.53}
& {78.7}& {88.8}& {18.56}
&  {78.7}&   {88.8}&{18.53}
&  {78.7}&   {88.8}&{18.53}\\

& COMBO~\cite{yang2024combo}
& {26.062}& {14.888}& {0.190} 
& {84.7}& {91.9}& {84.6}&    {91.9}&{19.00}
& {84.6}& {91.9}& {19.00}
&  {84.6}&   {91.9}&{19.00}
& {84.6}& {91.9}&{18.9}\\

% \rowcolor{lightgray} 
\cellcolor{white} & {TPVAI + CL~\cite{li2025reallyavs}} 
& {87.672}& {82.461}& {0.000} 
& {78.1}& {88.2} & {0.2}& {22.6}& {0.00}
& {0.2}& {22.6}& {0.00}
& {0.2}& {22.6}& {0.00}
& {0.61}&  {17.31}& {0.00}\\

% \rowcolor{lightgray}     
\cellcolor{white} & {AVSegFormer + CL~\cite{li2025reallyavs}} 
& {85.069}& {80.849}& {0.001} 
& {74.2}& {84.8} & {0.2}& {22.6}& {0.00}
& {0.3}& {22.7}& {0.03}
& {0.5}& {22.9}& {0.10}
& {0.16}&  {22.59}& {0.00}\\ 

\midrule

\multirow{7}{*}{{AVSBench-MS3}}

& TPVAI~\cite{zhou2024avss}
& {59.468}& {51.036}&{0.072}
& {54.0}& {64.5}& {27.6}&    {53.5}&{5.23}
& {31.7}& {57.4}& {5.37}
&  {42.2}&   {62.4}&{9.49}
&  {42.24}&  {62.16}&{9.53}\\

& AVSegFormer~\cite{gao2024avsegformer}
& {54.889}& {46.571}&{0.103}
& {61.3}& {73.8}& {53.2}&    {68.2}&{12.73}
& {47.5}& {63.8}& {9.23}
&  {50.3}&   {66.0}&{10.74}
& 51.14& 66.24&11.73\\

& Stepping-Stones~\cite{ma2024stepping}
& {61.439}& {43.937}&{0.114}
& {67.3}& {77.6}& {45.6}&  {72.5}&{8.93}
& {43.8}& {72.3}& {8.08}
&  {41.0}&   {63.3}&{15.36}
& 45.81& 68.04&12.48\\

& SAMA-AVS~\cite{liu2024SAMA-AVS-Syntheticdataset}
& {65.308}& {65.038}&{0.125}
& {68.6}& {78.3}& {29.7}&  {36.8}&{9.22}
& {39.2}& {46.6}& {11.79}
&  {44.1}&   {49.9}&{13.96}
& {41.9}&  {47.5}&{14.36}\\

& CAVP~\cite{chen2024cavp-VPO}
& {49.647}& {47.262}&{0.110}
& {45.8}& {61.7}& {45.8}&    {61.7}&{11.01}
& {45.8}& {61.7}& {11.06}
& {45.8}&   {61.7}&{11.02}
& {45.81}&  {61.72}&{11.02}\\

% \rowcolor{lightgray} 
\cellcolor{white} &  {TPVAI + CL~\cite{li2025reallyavs}} 
&    {65.427}&  {70.911}& {0.001}
& {51.3}&  {64.5}&  {9.8}&  {17.7}&  {0.42}
&  {9.9}&  {25.8}&  0.19
&  {9.1}&  {20.3}&  {0.04}
& {9.06}& {33.96}&{0.82}\\ 

% \rowcolor{lightgray} 
\cellcolor{white} &  {AVSegFormer + CL~\cite{li2025reallyavs}} 
& {73.354}&  {78.244} & {0.000}
& {61.5}&   {74.0}&  {9.1}&  {17.0}&  {0.00}
&  {9.4}&  {17.2}&  {0.00}
&  {9.1}&  {17.0}&  {0.00} 
& {9.06}& {16.96}&  {0.00}\\ 
    
\midrule
\multirow{6}{*}{{AVSBench-Semantic}}

& TPVAI~\cite{zhou2024avss} 
& {41.895}& {45.511}& {0.103}
& {29.8}& {35.2}& {29.3}& {35.8}& {10.07}   
& {30.2}& {36.1}& {10.54}
& {28.9}& {35.0}& {10.02}
& {19.2}& {23.0}& {24.06}\\ 

& AVSegformer~\cite{gao2024avsegformer} 
& {47.409}& {49.689}& {0.116}
& {37.4}& {42.8}& {35.5}& {41.2}& {11.81}
& {35.4}& {40.8}& {11.93}
& {34.4}& {40.2}& {11.37}
& {35.7}& {41.4}& {11.95}\\

& CAVP~\cite{chen2024cavp-VPO} 
& {24.522}& {34.987}& {0.034}
& {14.1}& {21.8}& {5.8}& {9.1}& {2.43}
& {7.5}& {11.4}& {2.00}
& {8.7}& {13.9}& {5.00}
& {8.15}& {12.76}&{3.32}\\
 
 & Stepping- Stones~\cite{ma2024stepping} 
& {49.099}& {48.030}& {0.184}
& {52.63}& {57.14}& {54.95}& {59.57}& {18.38}
& {54.39}& {59.01}& {18.38}
& {52.63}& {57.14}& {18.37}
& {54.29}& {58.90}&{18.38}\\ 

% \rowcolor{lightgray} 
\cellcolor{white} & TPVAI + CL~\cite{li2025reallyavs}
& {42.303}& {50.050}& {0.006}
& {27.0}& {33.6}& {1.20}& {1.26}& {5.30}
& {1.23}& {1.28}& {0.10}
& {4.64}& {2.26}& {0.17}
& {2.38}& {1.40}& {0.13}\\ 

% \rowcolor{lightgray} 
\cellcolor{white} & AVSegformer + CL~\cite{li2025reallyavs}
& {54.617}& {65.092}& {0.003}
& {37.6}& {48.4}& {0.0}& {0.0}& {0.01}
& {0.1}& {0.1}& {0.05}
& {1.3}& {1.6}& {0.01}
& {0.00} & {0.01} & {0.01}\\ \hline
\end{tabular}%
}

\vspace{-5mm}

\end{table*}

\paragraph{Performance on Robust-AVSBench~\cite{li2025reallyavs}}
\textbf{\textit{Most AVS models fail to leverage audio cues effectively, exposing a strong visual bias.}}
Evaluations on Robust-AVSBench (Table~\ref{tab:robustAVS}) reveal that many otherwise strong models, such as Stepping-Stones~\cite{ma2024stepping} and CAVP~\cite{chen2024cavp-VPO}, are not robust to challenging audio conditions. Critically, their segmentation performance remains nearly identical across silent, irrelevant, and noisy audio inputs. This performance, coupled with high False Positive Rates and consequently low global metrics (G-mIoU), indicates that these models largely ignore the audio modality and rely on a strong visual bias. While this issue is most pronounced on the S4 benchmark, it persists across MS3 and AVSS as well.

\vspace{-2mm}
\section{Challenges and Future Directions}
\label{sec:challenges_future}

\paragraph{Challenges}
The analysis presented in Section~\ref{sec:results} reveals several critical limitations in current AVS models. 
% Current AVS models face several critical limitations. 
Insufficient temporal modeling significantly hampers their ability to accurately localize sound-emitting objects in long, untrimmed video sequences, as highlighted by benchmarks like LU-AVS~\cite{liu2024LU-AVS}. Additionally, a pronounced modality bias, characterized by an excessive reliance on visual information and inadequate integration of audio cues, is evident, particularly under negative audio conditions in Robust-AVSBench evaluations~\cite{li2025reallyavs}. Furthermore, AVS systems often struggle in real-world scenarios due to background noise, occlusions, visually similar distractors, and small-scale or deformable targets. Finally, computational inefficiency remains a significant barrier, restricting the deployment of resource-intensive state-of-the-art models~\cite{gao2024avsegformer, chen2024cavp-VPO, ma2024stepping} in real-time or on resource-constrained devices.

\paragraph{Future Directions Discussion}

Future research should prioritize enhanced temporal reasoning using temporal attention mechanisms and sequential modeling frameworks~\cite{li2018unified, li2020spatio, yan2019stat, guo2022attention}. Effective multimodal fusion techniques, such as contrastive learning, gradient modulation, and conflict-resolving gradient integration~\cite{khosla2020supervised, chen2020simple, peng2022balanced, wei2024mmpareto}, are necessary to mitigate modality imbalance. Additionally, methods improving robustness against noise, occlusions, and visually similar distractors through realistic training protocols and data augmentation are critical. Advancing computational efficiency via lightweight architectures, model compression, quantization, and knowledge distillation~\cite{han2015deep, hubara2018quantized, hinton2015distilling} will enable practical real-time deployments. Exploring weakly supervised and unsupervised approaches leveraging natural audio-visual co-occurrences~\cite{arandjelovic2017look, mo2023weakly-supervised, korbar2018cooperative} can reduce annotation costs significantly. Finally, integrating large multimodal models (LMMs)~\cite{radford2021learning, li2023blip, achiam2023gpt} holds potential for enhancing generalization and cross-modal reasoning in AVS.

% ==================================================
% Section: Conclusion
% ==================================================
\vspace{-2mm}
\section{Conclusion}
\label{sec:conclusion}

AVS has emerged as a crucial task within multimodal learning, aiming to segment visual objects based on associated audio cues. This survey reviewed AVS development methodologies, including multimodal encoding-fusion-decoder architectures, training paradigms, benchmarks, and results. Our analysis highlighted significant advancements alongside persistent challenges such as weak temporal modeling, modality imbalance, limited robustness to real-world conditions, and computational inefficiency. Future research directions emphasize improving temporal understanding, balanced multimodal fusion, leveraging large multimodal foundation models, minimizing annotation costs through unsupervised techniques, and integrating high-level semantic reasoning to foster more intelligent AVS systems.

\vspace{-3mm}

\section*{Acknowledgments}

The authors used ChatGPT~\cite{achiam2023gpt} for minor grammar and language editing. All technical content, analysis, and writing were created by the authors.

\bibliographystyle{IEEEtran}
\bibliography{main}

\end{document}